\documentclass[lettersize,journal]{IEEEtran}
\usepackage{amsmath,amsfonts}
\usepackage{algorithmic}
\usepackage{array}
\usepackage[caption=false,font=normalsize,labelfont=sf,textfont=sf]{subfig}
\usepackage{textcomp}
\usepackage{stfloats}
\usepackage{url}
\usepackage{verbatim}
\usepackage{graphicx}
\usepackage{pgf-pie}
\usepackage{tikz}
\usepackage{tikzscale}
\usepackage{tkz-graph}
\usetikzlibrary{shapes,backgrounds,patterns,arrows}
\usepackage{pgfplots}
\def\BibTeX{{\rm B\kern-.05em{\sc i\kern-.025em b}\kern-.08em
    T\kern-.1667em\lower.7ex\hbox{E}\kern-.125emX}}
\usepackage{balance}

\usepackage{xurl}
\usepackage{xcolor}
\usepackage{svg}

\usepackage{multirow}

\definecolor{orange}{rgb}{1,0.8,0}
\definecolor{gray}{rgb}{.9,0.9,0.9}
\definecolor{darkgray}{rgb}{.3,0.3,0.3}
\definecolor{darkblue}{rgb}{.1,0.0,0.3}
\definecolor{lightblue}{rgb}{0.7,0.7,1}
\definecolor{lightred}{rgb}{1,0.7,.7}
\definecolor{purple}{RGB}{204,153,255}
\definecolor{lightgray}{rgb}{.95,0.95,0.95}
\definecolor{lightgreen}{rgb}{0.3,0.5,0.3}
\definecolor{darkgreen}{rgb}{0.05,0.3,0.05}
\definecolor{rosso}{RGB}{220,57,18}
\definecolor{giallo}{RGB}{255,153,0}
\definecolor{blu}{RGB}{102,140,217}
\definecolor{verde}{RGB}{16,150,24}
\definecolor{viola}{RGB}{153,0,153}

\def\editmode{0}

\if\editmode1   % EDIT MODE 
    \newcommand{\acom}[1]{\noindent\textcolor{red}{{[#1]}}} % author's comment
    \newcommand{\rch}[1]{\noindent\textcolor{red}{\sout{#1}}} % track (removed) changes in the document
    
    \newcommand{\cmt}[1]{\leavevmode\linebreak\noindent\textcolor{lightgreen}{\underline{[#1]}}\linebreak} % top-level structural comment
    
    \newcommand{\scmt}[1]{\leavevmode\newline\indent$\bullet$\hspace{1pt}\textcolor{lightgreen}{\underline{[#1]}}} % first-level bullet comment
    
    \newcommand{\sscmt}[1]{\leavevmode\newline\indent\indent--\hspace{1pt}\textcolor{lightgreen}{\underline{[#1]}}} % second-level bullet comment 

\else   % VIEW MODE 
    \newcommand{\acom}[1]{\ignorespaces}   
    
    \newcommand{\rch}[1]{\ignorespaces}
    
    \newcommand{\cmt}[1]{\ignorespaces} 
    \newcommand{\scmt}[1]{\ignorespaces}
    \newcommand{\sscmt}[1]{\ignorespaces}

\fi

\tikzset{
  chart/.style={
    legend label/.style={font={\scriptsize},anchor=west,align=left},
    legend box/.style={rectangle, draw, minimum size=5pt},
    axis/.style={black,semithick,->},
    axis label/.style={anchor=east,font={\tiny}},
  },
  pie chart/.style={
    chart,
    slice/.style={line cap=round, line join=round, very thick,draw=white},
    pie title/.style={font={\bfseries}},
    slice type/.style 2 args={
        ##1/.style={fill=##2},
        values of ##1/.style={}
    }
  }
}

\pgfdeclarelayer{background}
\pgfdeclarelayer{foreground}
\pgfsetlayers{background,main,foreground}

\newcommand{\pie}[3][]{
    \begin{scope}[#1]
    \pgfmathsetmacro{\curA}{90}
    \pgfmathsetmacro{\radius}{1}
    \def\Centre{(0,0)}
    \node[pie title] at (90:1.3) {#2};
    \foreach \v/\s in{#3}{
        \pgfmathsetmacro{\deltaA}{\v/100*360}
        \pgfmathsetmacro{\nextA}{\curA + \deltaA}
        \pgfmathsetmacro{\midA}{(\curA+\nextA)/2}

        \path[slice,\s] \Centre
            -- +(\curA:\radius)
            arc (\curA:\nextA:\radius)
            -- cycle;

   \pgfmathsetmacro{\ysign}{ifthenelse(mod(\midA,360)<=180,1,-1)}
   \pgfmathsetmacro{\xsign}{ifthenelse(mod(\midA-90,360)<=180,-1,1)}

   \begin{pgfonlayer}{foreground}
        \draw[*-,thin] \Centre ++(\midA:\radius/2) -- 
                               ++(\xsign*0.07*\radius,\ysign*0.2*\radius) -- 
                               ++(\xsign*\radius,0) 
                      node[above,near end,pie values,values of \s]{$\v\%$};
   \end{pgfonlayer}

        \global\let\curA\nextA
    }
    \end{scope}
}

\newcommand{\legend}[2][]{
    \begin{scope}[#1]
    \path
        \foreach \n/\s in {#2}
            {
                  ++(0,-10pt) node[\s,legend box] {} +(5pt,0) node[legend label] {\n}
            }
    ;
    \end{scope}
}

\title{Interplay between Federated Learning and Explainable Artificial Intelligence: a Scoping Review
\thanks{
The work in this paper was supported by the VALIDATE project grant 101057263 from the EU HORIZON-RIA.}
}

\author{Luis M. Lopez-Ramos$^{1}$, Florian Leiser$^{2}$, Aditya Rastogi$^{3}$, Steven Hicks$^{4}$, \\ Inga Strümke$^{5}$, Vince I. Madai$^{6,7}$, Tobias Budig$^{8}$, Ali Sunyaev$^{9}$, and Adam Hilbert$^{10}$ \\ On behalf of the VALIDATE consortium
    \thanks{$^{1}$ \texttt{luis@simula.no}, Holistic Systems department, Simula Metropolitan Center for Digital Engineering, Oslo, Norway.}
    \thanks{$^2$\texttt{florian.leiser@kit.edu}, Institute of Applied Informatics and Formal Description Methods, Karlsruhe Institute of Technology, Germany.}
    \thanks{$^3$\texttt{aditya.rastogi@ukbonn.de}, Department of Neuroradiology, University Hospital Bonn, Germany.}
    \thanks{$^4$\texttt{steven@simula.no}, Holistic Systems department, Simula Metropolitan Center for Digital Engineering, Oslo, Norway.}
    \thanks{$^5$\texttt{inga.strumke@ntnu.no}, Department of Computer Science, Faculty of Informatics, Norwegian University of Science and Technology, Trondheim, Norway.}
    \thanks{$^{6}$\texttt{vince\_istvan.madai@bih-charite.de}, QUEST Center for Responsible Research, Charité - Universitätsmedizin Berlin, Berlin, Germany.}
    \thanks{$^{7}$School of Computing and Digital Technology, Birmingham City University, Birmingham, United Kingdom.}
    \thanks{$^{8}$\texttt{tbudig@student.ethz.ch}, ETH Zurich, Switzerland}
    \thanks{$^9$\texttt{ali.sunyaev@tum.de}, School of Computation, Information and Technology, Technical University of Munich (Campus Heilbronn), Germany.}
    \thanks{$^{10}$\texttt{adam.hilbert@charite.de}, Charité Lab for AI in Medicine (CLAIM), Charité - Universitätsmedizin Berlin, Germany.}}

\begin{document}

\maketitle

\begin{abstract}

The joint implementation of federated learning (FL) and explainable artificial intelligence (XAI) could allow training models from distributed data and explaining their inner workings while preserving essential aspects of privacy. Toward establishing the benefits and tensions associated with their interplay, this scoping review maps the publications that jointly deal with FL and XAI, focusing on publications that reported an interplay between FL and model interpretability or post-hoc explanations.
Out of the 37 studies meeting our criteria, only one explicitly and quantitatively analyzed the influence of FL on model explanations, revealing a significant research gap. The aggregation of interpretability metrics across FL nodes created generalized global insights at the expense of node-specific patterns being diluted. Several studies proposed FL algorithms incorporating explanation methods to safeguard the learning process against defaulting or malicious nodes. 
Studies using established FL libraries or following reporting guidelines are a minority. More quantitative research and structured, transparent practices are needed to fully understand their mutual impact and under which conditions it happens.
%more papers focusing on explanation methods (mainly feature relevance) than on interpretability (primarily algorithmic transparency).
%Most works used simulated horizontal FL setups involving 10 or fewer data centers.

\end{abstract}

\section{Introduction}
\label{s:introduction}

% Artificial Intelligence (AI) systems are increasingly relied upon in high-stakes domains such as healthcare, finance, and engineering, where both privacy preservation and model comprehension are essential. On the one hand, users provide their data only if sufficient privacy is granted. Federated learning (FL) has gained prominence as a solution to train models from distributed data while preserving these privacy aspects. On the other hand, model comprehension is essential for users to rely on model outputs. Explainable artificial intelligence (XAI) aims to explain the inner workings of machine learning (ML) algorithms, thereby increasing model comprehension. 
% Despite the significance of both, the potential interaction between these technologies remains unclear. FL and XAI could be implemented independently and provide both benefits. Alternatively, their interplay could create tensions and tradeoffs between comprehension and privacy.
% %To answer this questions, exploratory research on the potential mutual influence between FL and XAI as enabling technologies is necessary.
% To address this knowledge gap, this paper presents a scoping review that examines the existing literature on the intersection of FL and XAI, exploring both, potential conflicts and opportunities for mutually reinforcing synergies.

% \input{figures/venn-diagram}

\begin{figure}
    \centering
    \includegraphics[width=0.99\columnwidth]{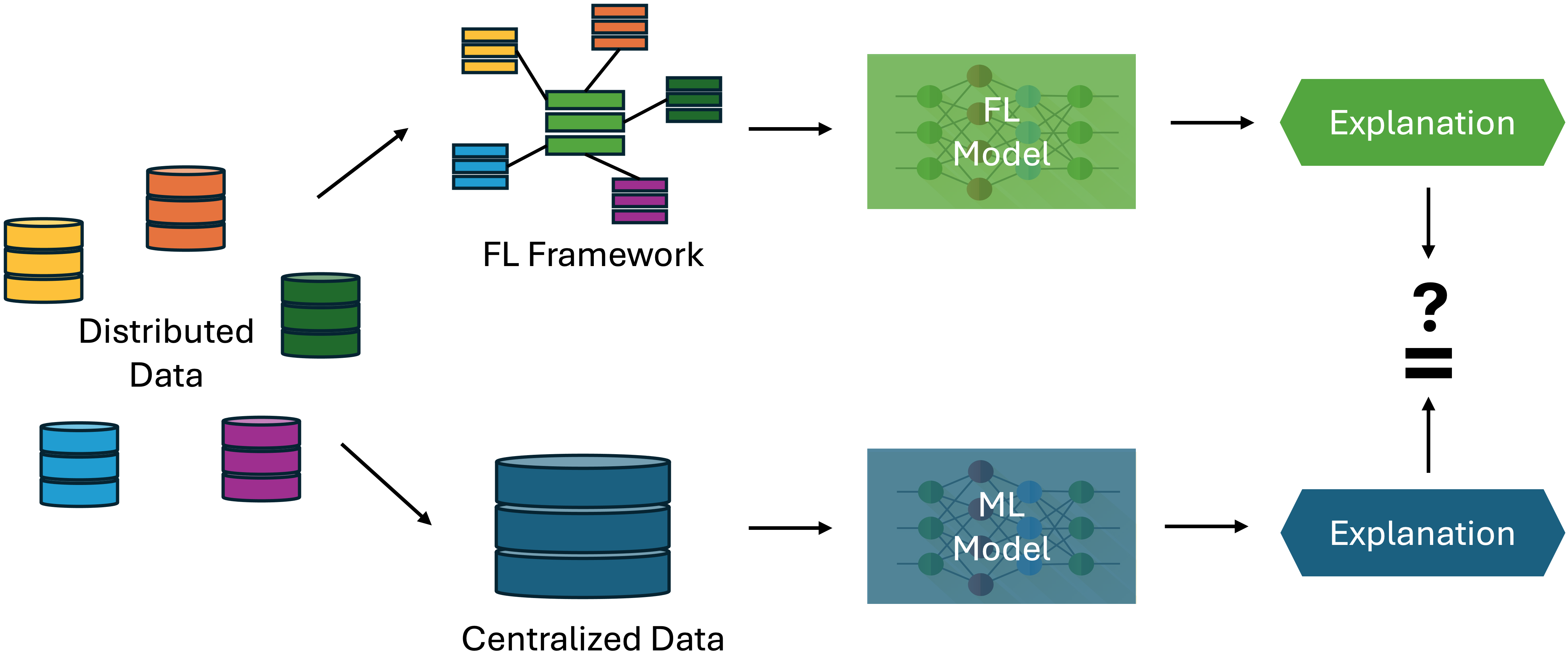}
    %\vspace{-1.5cm}
    \caption{Schematic overview of a main research question: do explanations differ between federated models and centralized models?}
    \label{fig:schema}
\end{figure}

The development of trustworthy AI systems depends on different ethical principles like privacy preservation and explicability~\cite{ethics_guidelines,EUcomission_2019}. Upholding these principles is essential to foster trust, ensure compliance, and maintain ethical integrity, particularly in data-sensitive fields like banking~\cite{doguc2022data} and healthcare~\cite{rieke_future_2020,saraswat2022explainable}. 

Privacy concerns often limit AI models in healthcare to data from single institutions. Anonymization of large-scale healthcare data is difficult due to the risk of patient re-identification~\cite{branson_evaluating_2020}. Federated learning (FL) helps solve this problem by allowing model training across multiple institutions without sharing raw data~\cite{yang2020FLbook,bak2024federated}. FL enables machine learning (ML) models to train on distributed data while protecting privacy. It also addresses governance concerns by restricting direct access to sensitive information~\cite{liu2021learning,mcmahan2017communication}. Training on data from multiple institutions improves model generalizability and reduces bias compared to models trained on homogeneous datasets~\cite{kolobkov2024efficacy}.

FL is categorized into three main types: horizontal FL (HFL), vertical FL (VFL), and federated transfer learning (FTL). HFL involves multiple clients holding different subsets of data points with the same feature space. VFL applies when institutions share data samples but with different feature spaces. FTL facilitates knowledge transfer across different FL settings, adapting models to new environments~\cite{yang2019federated}.

Explicability is another key ethical principle emphasized by ethicists and policymakers in high-risk AI domains like healthcare~\cite{EUcomission_2019}. It requires that (a) AI processes be transparent, (b) the capabilities and purpose of AI systems be clearly communicated, and (c) AI decisions and predictions be explainable to those directly or indirectly affected~\cite{ethics_guidelines,EUcomission_2019}. AI models that are transparent and understandable are more likely to earn the trust and acceptance of stakeholders, including customers, regulators, and users~\cite{markus_role_2021}. According to European Union regulations, aspects (a) and (c) of explicability, technically referred to as explainability~\cite{herzog2022risk}, must be supported by additional information on how an AI system arrives at its outputs beyond performance metrics alone~\cite{amann_explain_2022}. Explainable artificial intelligence (XAI) focuses on developing methods that make ML models more understandable. This can be achieved through interpretable model architectures or post-hoc explanation methods, aligning with requirements (a) and (c) of the EU guidelines~\cite{linardatos2020explainable}.

However, the XAI literature lacks a consensus on the definition of explainability. Terms like \emph{explainability}, \emph{explicability}, and \emph{interpretability} are often used interchangeably~\cite{adadi_peeking_2018}. Some distinctions exist between interpretable modeling and explanation methods~\cite{amann_explain_2022}, which we adopt here due to their growing importance in high-stakes AI applications like healthcare~\cite{rudin2019stop}. An ML model is considered \emph{interpretable} if its decision process can be inherently and intuitively understood by the intended user~\cite{amann_explain_2022}. The term inherently is crucial, as interpretability is a \emph{passive} property that reflects how naturally a model's decisions make sense to a human observer~\cite{barredo_arrieta_explainable_2020} without requiring additional computation. Examples include linear and logistic regression models, where variable importance is inferred from model weights, and decision trees, which humans can intuitively follow~\cite{amann_explain_2022}. 
In contrast, explanation methods \emph{actively} perform additional computations on non-interpretable models to clarify their internal functions~\cite{barredo_arrieta_explainable_2020,amann_explain_2022}. These computations may involve the evaluation of model gradients for perturbed inputs~\cite{krishna2024post}, assessing feature importance~\cite{lundberg_unified_2017}, analyzing how output varies with feature changes~\cite{szepannek2023much}, or identifying minimal input changes that alter predictions~\cite{guidotti2022counterfactual}. Explanation methods serve as interfaces between humans and AI systems while approximating the decision-making process~\cite{guidotti_survey_2019}.

Extensive research has been conducted on FL and XAI separately~\cite{saeed2023explainable,daly2024federated}, but their combined role in trustworthy AI remains underexplored. It is unclear whether FL and XAI complement each other, impose conflicting requirements, or can be addressed independently. XAI can help mitigate risks in federated learning by addressing challenges associated with decentralized data. Learning from heterogeneous datasets can introduce spurious correlations, as individual subsets may contain biases that distort the representation of the overall data distribution. Explainability helps practitioners assess whether model predictions are based on valid patterns rather than misleading correlations, reducing unintended biases and ensuring ethical compliance. Additionally, explainability can aid in detecting malicious agents attempting to poison the FL process. However, the interplay between FL and XAI entails potential risks that should be thoroughly examined and documented. There is no unified effort to determine whether FL reduces interpretability or affects the accuracy of explanations. Furthermore, explanations could expose vulnerabilities in the FL network, increasing susceptibility to attacks. Preliminary research~\cite{hilbert2023effect} suggests that model explanations differ between FL and traditional centralized models, highlighting unique challenges in applying XAI in FL contexts. While many studies highlight the benefits of both perspectives, few quantitatively analyze their mutual impact. The literature also lacks a systematic examination of methodologies, challenges, and outcomes related to the interplay of FL and XAI, leaving a gap in understanding their interaction.

%Researchers interested in analyzing the aforementioned interplay need to be aware of the reported findings and the setups and conditions where these findings were obtained. To address this research gap, the present scoping review maps the relevant work, both experimental and methodological, focusing on the impact of one on the other, and analyzing the technical settings under which these research works have been carried out. More specifically, we look at publications that investigate the interaction of FL and XAI and either i) propose FL methods that include explainability in the algorithm design, %ii) discuss practical settings where FL and XAI are jointly applied, or iii) characterize the interplay between FL and explainability %(Sec. \ref{ss:assessing}) 
%employing analysis or experiments.

To understand the interplay between FL and XAI, it is crucial to consider the findings, setups, and conditions under which these studies have been conducted. This scoping review maps experimental as well as methodological studies that explore their interaction. Since FL can potentially influence model interpretability and post-hoc explanations, the review investigates both aspects of XAI.
% To cover the relevant literature it is necessary to search based on the words ``explanation'' and ``interpretability'' (see Table \ref{tab:search_terms}) as different authors use the words interchangeably.
We focus on research articles that either propose FL methods incorporating explainability, discuss practical applications where FL and XAI are jointly used, or analyze the interplay between these technologies through empirical studies. Given the fragmented and emerging nature of research at this intersection, this review provides a comprehensive overview of current work, identifies key concepts and evidence types, and highlights gaps to guide future research efforts.

The remainder of this paper is structured as follows. Section \ref{sec:related_work} introduces the related work and Sec. \ref{s:methodology} describes the method we followed. Analysis of the information extracted from the selected studies is presented in Sec. \ref{s:results_new}, and a more detailed analysis of the interplay between FL and XAI found in the reviewed papers is provided in Sec. \ref{s:interplay}. Section \ref{s:discussion} discusses the implications of the reported results before Sec. \ref{s:conclusion} concludes the paper. The Appendix provides a summary of how all included studies jointly address FL and XAI.

\section{Related Work}
\label{sec:related_work}

Review papers focusing separately on XAI and FL are abundant, and the literature even includes several meta-reviews \cite{saeed2023explainable} and systematic reviews from different points of view (see, e.g., \cite{crowson2022systematic, Li2023_Survey_FL} regarding FL applications for biomedical data). 
An extensive review of XAI methods can be found in \cite{linardatos2020explainable} or \cite{Li2022_Survey_XAI}. Although there are a few reviews concerning the joint application of XAI and FL, none analyzes the interplay in as much depth as outlined in the following sections.

\subsection{Reviews about FL and XAI}

%---Adam---
The review in~\cite{r2_barcena2022fed} introduces Fed-XAI, which involves both the learning of interpretable models within a federated setting and the application of explanation methods to any federated model. They use \cite{barredo_arrieta_explainable_2020} as a source for XAI taxonomy and highlight the diversity of definitions used in the studied literature. Additionally, they give an overview of the ``current status in Fed-XAI'' by discussing a set of articles relating to the two concepts with no analysis of the interplay between FL and XAI. 

The review in \cite{r1_lopezblanco2023federated} gathers relevant articles concerning the ``FED-XAI'' concept, defined by them as a discipline ``which aims to bring together these two approaches into one''. Differing from our search terms, which include variations of the words ``explanation'' and ``interpretability'', their search is limited to the explicit string ``FED-XAI'', returning a reduced number of publications. The manuscript lacks a comparative analysis of the interplay between FL and XAI among the found articles. 

The review in~\cite{r3_li2023towards} surveys ``interpretable federated learning'' and proposes an interpretable FL taxonomy that enables learning models to explain prediction results, support model debugging, and provide insights into data owner contributions. The survey analyzes representative interpretable FL approaches, commonly adopted performance evaluation metrics, and future research directions. The primary focus is on leveraging interpretable methods to improve the FL process, but not on the impact of FL on the explanations of the predictions.

In contrast to these works, our review 1) encompasses an ample set of articles through a scoping review methodology, 2) provides detailed results about how FL and XAI are applied in each study, 3) summarizes how the reviewed studies jointly deal with FL and XAI, and 4) analyzes the findings on different kinds of interplay between them.

\subsection{Contribution-Aware Federated Learning}
\label{ss:contribution_aware}

A subset of the approaches investigating the intersection of FL and XAI has utilized feature relevance methods (e.g., Shapley values) to measure the contribution of each client participating in an FL process. These approaches, commonly referred to as \textit{contribution-aware FL}, aim to incentivize clients to engage during model training and to distribute rewards fairly based on contributions. One of the first approaches to contribution-aware FL suggested using Shapley values to interpret contributions in FL networks in \cite{wang2019interpret}. The contributions are diverse, ranging from node liability \cite{p106_malandrino2021node} to biases within data sets \cite{pandl2023reward}. Recent approaches have also extended SHAP-based~\cite{lundberg_unified_2017} contribution determination to provide visualizations to evaluate data privacy \cite{Guo2023} and to improve prediction reliability via clustering patients \cite{p15_yuan2022efficient}. Such methods do not necessarily constitute an influence of XAI on the FL process and are, therefore, not central to the present review.

\section{Methods}
\label{s:methodology}

The present scoping review addresses the joint application of FL and XAI and their potential interplay. % Two research types have been addressed, namely the methodological advances and the experimental results. 
We aimed to examine the range and nature of research activity on the topic, summarize the research findings, and identify research gaps, following a standard scoping review methodology \cite{arksey2005scoping}. The presence of two concepts of explainability (interpretable models and post-hoc explanations), the two research types (methodological and experimental), and the diversity of joint approaches to FL and XAI increased the complexity of the study. Understanding and summarizing the diverse approaches required a deeper analysis of each included article.
A more detailed description of the methodology is available in the pre-published research protocol \cite{protocol}.

\subsection{Identifying Research Questions}
\label{RQs}
The research questions of this work address the training of interpretable ML models using FL and the explanation methods applied to ML models trained via FL. Furthermore, we categorized the research questions into two groups:

\subsubsection{Methodological Advances}
\begin{itemize}
    \item Which interpretable ML models can be trained via FL? 
    \item What are existing methods to train interpretable ML models via FL?  
    \item Are there explanation methods that take into account that the ML model was trained via FL? 
    \item Has any study proposed an FL method that takes into account the ulterior application of explanation methods?
    \item For what subtypes of FL have methods been proposed to a) learn interpretable ML models; b) explain the outputs of ML models?
\end{itemize}

\subsubsection{Experimental Results}
\begin{itemize}
    \item In which contexts (fields of application) have FL and XAI been jointly evaluated?
    \item Under what conditions does training a model via FL affect a) its interpretability? b) the obtained explanations?
    \item Has any study compared the effects of FL on the interpretability of the resulting models against a centralized learning algorithm? 
    \item Has any study compared the explanations obtained from a model trained via FL against a centralized learning algorithm?
\end{itemize}

\subsection {Identifying Relevant Studies}
We tried to gather a broad range of manuscripts. Therefore, we used the five databases IEEEXplore, Google Scholar, PubMed, Scopus, and Web of Science. The terminological ambiguity between explainability and interpretability influenced the search strategy, which aimed to identify as many articles as possible that mention FL in conjunction with the (broad) concept of explainability. The search terms are detailed in Table \ref{tab:search_terms}.
\begin{table}[ht]
    \centering
    \begin{tabular}{|p{2cm}|p{\dimexpr\linewidth-2cm-4\tabcolsep-2\arrayrulewidth}|}
        \hline
        \textbf{Library/Engine} & \textbf{Search string} \\
        \hline
        IEEEXplore & (``federated learning'') AND ((explainable) OR (interpretable) OR (explaining) OR (explainability) OR (interpreting) OR (interpretability) OR (interpret)) \\
        \hline
        Google Scholar & allintitle: (explainable OR explainability OR explaining OR interpretability OR interpretable OR interpret OR intepreting)  federated learning \\
        \hline
        PubMed & (``federated learning'') AND (explainable OR explaining OR explainability OR interpret OR interpreting OR interpretable OR interpretability) \\
        \hline
        Scopus & TITLE-ABS(\{federated learning\} AND (explainable OR explaining OR explainability OR interpret OR interpreting OR interpretable OR interpretability)) AND (LIMIT-TO(DOCTYPE, ``ar'') OR LIMIT-TO(DOCTYPE, ``cp''))\\
        \hline
        Web Of Science & Federated Learning AND (explainable OR explaining OR explainability OR interpret OR interpreting OR interpretable OR interpretability) \\
        \hline
    \end{tabular}
    \vspace{2mm}
    \caption{Search terms}
    \vspace{-5mm}
    \label{tab:search_terms}
\end{table}

\subsection{Study Selection}

\begin{figure}
    \centering
    \vspace{-6mm}
    \includegraphics[width=1.4\columnwidth]{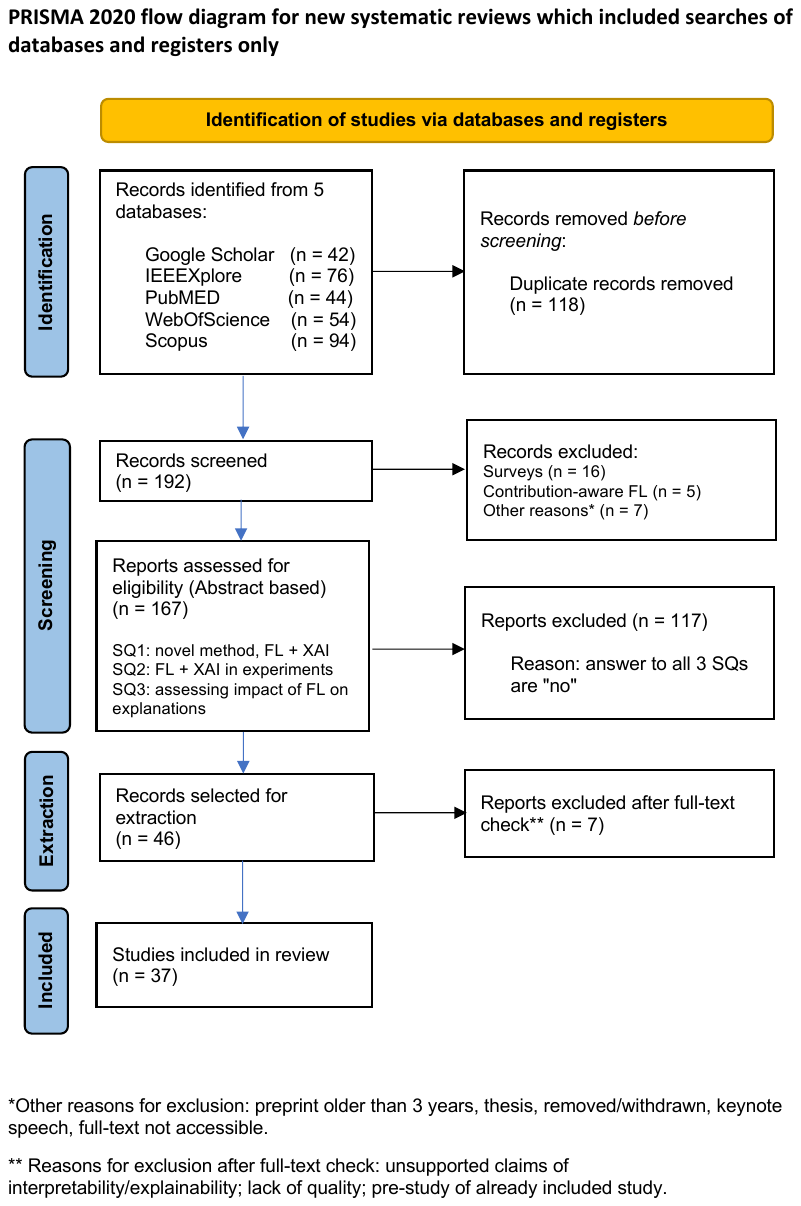}
    \vspace{-5cm}
    \caption{Flow diagram indicating the number of  papers identified, excluded in the different phases, and included in the review.}
    \label{fig:prisma}
    
\end{figure}

We initially identified research articles published in peer-reviewed conferences and journals and works made available in pre-print services such as arXiv over the past three years. We excluded theses, reviews, meta-reviews, and surveys since they aim to summarize existing efforts and are discussed above. In our survey, the time frame of the published articles was between 2019 and April 2023 (date of protocol pre-publication). No relevant studies were found before 2019. We required that the full text be available. A flowchart summarizing the number of articles found, excluded, and included in this study, along the lines of \cite{page2021prisma}, is displayed in Fig. \ref{fig:prisma}.

To be included in the review, research articles had to discuss the relation and interaction between XAI and FL or apply both technologies jointly in a practical setting. We performed a screening procedure to rule out papers that only mention both technologies without detailed discussion or practical application. To that end, articles must answer positively to at least one of the following screening questions: 
\begin{itemize}
    \item SQ1) Does the article propose a novel method integrating FL and XAI?
    \item SQ2) Does the article report results from experiments applying FL and XAI in a real dataset?
    \item SQ3) Does the article discuss or assess the impact of FL on explanations or model interpretability?
\end{itemize}

\subsection{Charting the Data}

We utilized a shared online spreadsheet to facilitate data extraction. During both the screening and extraction process, double coding was practiced with the involvement of all authors. Each screened paper was independently assessed by two authors, and the extraction points for each included study were likewise completed independently by two authors. We tried resolving differences in data extraction through discussion between the two involved authors. If the disagreements persisted, a discussion within the entire author consortium resulted in the final decision. While extracting, we distinguished between explanation methods and model interpretability according to the nomenclature introduced above, even though some articles did not agree with the latter. We collected the following data:

\begin{itemize}
    \item \textbf{Study identification:} title, authors, publication year, publication outlet.  
    \item \textbf{Nomenclature:} whether the article provides or cites a definition of explainability, and whether the nomenclature regarding explainability coincides with the one introduced here.
    \item \textbf{Nature of the study:} whether it is general, theoretical, or applied; if applied, what field it is applied on, and whether the proposed idea is practically validated in the field of application.
    \item \textbf{Data characteristics:} information modality, type, and amount (number of unique data samples) of the data used in the experiments.
    \item \textbf{FL-specific characteristics:} type of FL (HFL, VFL, FTL) used, setup (simulated or not, number of data centers), and which FL library is used (e.g., FLWR, openFL).
    \item \textbf{XAI-specific characteristics:} whether an interpretable model or an explanation method is used\footnote{If the nomenclature in an article does not agree with the one mentioned in the Introduction, we identify whether it uses explanation methods or interpretable models according to our understanding.}, its type, and the specific explanation method (e.g.: SHAP, GradCAM~\cite{selvaraju_grad-cam_2020}) or way of interpreting the model (e.g., weights at the first layer of DNN).
    \item \textbf{Interplay between FL and XAI}: a paragraph summarizing how the article deals jointly with FL and XAI; influence of FL on explanations or interpretability (e.g. significant differences in variable importance ratings between centrally learned and model trained via FL); influence of XAI onto federated training (e.g., modified merging step in the central server after collecting locally trained models); and whether this influence is quantified, and how.
    \item \textbf{Methodology notes:} how the novel method is designed (e.g., optimization-based approach), and how rigorous the methodology is.
\end{itemize}

\section{Results}
\label{s:results_new}

In this section, the information extracted from the 37 articles included in the review is collated, summarized, and reported. 
First, we give an overview of the studies where FL and XAI are applied together before presenting a more detailed description of the 11 articles where an interplay between FL and XAI was reported. Summaries of how the articles selected for review jointly deal with FL and XAI are provided in Appendix \ref{appendix:summaries}.

\cmt{Main findings regarding general sample of articles}

\begin{figure}[!t]
    \centering
    \includegraphics[width=\columnwidth]{tikz/pie-chart-explanation-methods}
    \caption{Distribution of reviewed articles across different XAI techniques. Explanation methods (shades of blue) were studied by 64.8\% of the included works, whereas interpretable models (shades of green) were studied by  35.1\% of the included works.}\label{fig:explainability}
\end{figure}

% \begin{figure}
%     \centering
%     \includegraphics[width=\columnwidth]{images/Explainability Methods.png}
%     \caption{Distribution of reviewed papers across different XAI techniques. Explanation methods (shades of blue) were studied by 64,8\% of the included works, whereas interpretable models (shades of green) were studied by  35,2\% of the included works.}
%     \label{fig:explainability}
% \end{figure}

\begin{table*}
    \centering
    \begin{tabular}{|l|l|l|} \hline 
           \textbf{Technique}& \textbf{Description} &\textbf{Manifestations in sample}\\ \hline 
           Feature relevance& Calculate relevance scores for model variables. &e.g., SHAP \cite{p15_yuan2022efficient, p257_fiosina2021interpretable, p16_demertzis2022explainable}, 
           \\  & &
           Custom builds~\cite{p1_younis2023flames, p243_xu2023federated}%, LIME\cite{p257_fiosina2021interpretable,p206_bensaad2022trust}
           \\\hline 
           Local explanations& Estimate whole model through less complex subsystems. &e.g., GradCAM \cite{p224_slazyk2022deep, p84_hou2021mitigating, p212_li2023privacy}, 
           \\  & &
           heatmaps \cite{p205rahman2021secure}\\ \hline 
           Simplification& Facilitate model while maintaining performance. &e.g., simpler models~\cite{p25_polato2022boosting, p14_barcena2022approach}\\ \hline 
           Text explanations& Generate symbols that explain the results of the model. &-\\ \hline 
  Visual explanations&Visualize the inference process of the model. &-\\ \hline 
  Explanations by example&Provide representative examples that allow insight into the model. &-\\ \hline 
           Algorithmic transparency& Enable users to follow and understand the processes by the model. &e.g., Linear regression \cite{p5_zheng2021federated}, 
           \\  & &
           Decision trees~\cite{p256_imakura2021interpretable, p48_chen2021fed}\\ \hline 
           Decomposability& Explain each model part separately for full comprehension. &e.g., Inference splits~\cite{p65_liang2022fedtsc}\\ \hline 
           Simulatability& The inference of models could be simulated by a human. &e.g., Rule-based systems~\cite{p33_pedrycz2021interpretability}\\ \hline
    \end{tabular}
    \vspace{4mm}
    \caption{Brief descriptions of the explainability techniques used in this review as understood in \cite{barredo_arrieta_explainable_2020}.}
    \label{tab:technique_ddescriptions}
\end{table*}

The collected interpretability and explanation methods are described in Table \ref{tab:technique_ddescriptions}. We see a slight tendency to the use of explanation methods over interpretability methods in our sample, as shown in Fig. \ref{fig:explainability}. Among the explanation methods applied with FL models, feature relevance is predominant, followed by local explanations, and a smaller amount of model simplification-based methods \cite{barredo_arrieta_explainable_2020}. Feature relevance methods quantify the impact of the model's input features, local explanations explain specific model predictions, and simplification refers to rebuilding the entire model for easier explainability. In the sample of selected articles, the most frequently used type of interpretability of FL models is algorithmic transparency, with some use of decomposability and simulatability \cite{barredo_arrieta_explainable_2020}. While algorithmic transparency allows users to algorithmically trace a model's processes, decomposability describes models of which fractions can be understood by humans. Simulatability refers to models that are sufficiently interpretable for a human to understand, or ``simulate'', as a whole \cite{barredo_arrieta_explainable_2020}. 
%---Florian---

\def\firstcircle{(0,0) circle (2cm)}
\def\secondcircle{(55:2.5
cm) circle (2cm)}
\def\thirdcircle{(0:3cm) circle (2cm)}

\begin{figure}[!t]
    \centering
    \includegraphics{tikz/venn-diagram-fl-type}
    \caption{Venn diagram showing the number of studies using each type of FL. Vertical and transfer FL are used far less than horizontal FL.}\label{fig:FL_type}
\end{figure}

% \begin{figure}
%     \centering
%     \includegraphics[width=0.7\columnwidth]{images/Venndiagream(1).png}
%     \caption{Venn diagram showing the number of papers using each type of FL. Vertical and transfer FL are used far less than horizontal FL.}
%     \label{fig:FL_type}
% \end{figure}

\if0
\begin{table}[h]
    \centering
    \begin{tabular}{|c|c|c|c|c|}
        \hline
        \textbf{Agreement} & \textbf{yes} & \textbf{no} & \textbf{-} & \textbf{SUM} \\ \hline
        \textbf{Definition} &  &  &  &  \\ \hline
        \textbf{yes} & 10 & 7 & 0 & 17 \\ \hline
        \textbf{no} & 7 & 4 & 9 & 20 \\ \hline
        \textbf{SUM} & 17 & 11 & 9 & 37 \\ \hline
    \end{tabular}
    \vspace{2mm}
    \caption{Nomenclature and definitions.}\label{t:nomenclature}
\end{table} %Florian
\fi

\begin{table}[h]
    \centering
    \begin{tabular}{|c|c|c|}
        \hline
        \textbf{Definition of XAI} & \textbf{XAI Nomenclature} & \textbf{Count} \\
        \hline
        \multirow{2}{*}{Provided} & Agrees with our protocol & 10 \\
        \cline{2-3}
        & Does not agree with our protocol & 7 \\
        \hline
        \multirow{3}{*}{Not provided} & Agrees with our protocol & 7 \\
        \cline{2-3}
        & Does not agree with our protocol & 4 \\
        \cline{2-3}
        & Unclear & 9 \\
        \hline
    \end{tabular}
    \vspace{2mm}
    \caption{Definition of explainability and nomenclature agreement}
    \label{tab:xai_definition}
\end{table}

Regarding the consistency of XAI nomenclature in the literature, we observed that only 17 out of 37 articles (49.46\%) provide a definition of XAI, out of which 10 agreed with our notion and 7 did not. When a definition was not given, 7 articles managed the term in accordance with our definition, 4 articles were in disagreement, and in 9 cases, it was not possible to infer which notion of XAI was used. Table \ref{tab:xai_definition} summarizes these findings.

%% FL %%

The most prominent type of FL used is horizontal, with 30 studies (81.1\%) using this type exclusively. VFL is used far less, appearing in only four articles, and TFL is the least used, with only one study focusing on it. The overlaps in the diagram show that not many articles used multiple of these methods. Only one article used horizontal and transfer FL, another used vertical and transfer FL. Fig. \ref{fig:FL_type} illustrates the distribution of FL types used in the reviewed studies.

\begin{figure}[!t]
    \centering
    \includegraphics[width=\columnwidth]{tikz/pie-chart-fl-libraries}
    \caption{Distribution of reviewed studies according to whether a FL library or an own implementation is used. Less than a third of the reviewed papers used established papers, and about a third did not specify how they implemented FL.}\label{fig:fl_libraries}
\end{figure}

% \begin{figure}
%     \centering
%     \includegraphics[width=\columnwidth]{images/Library Use (1).png}
%     \caption{Distribution of reviewed papers according to whether a FL library or an own implementation is used. Less than a third of the reviewed papers used established papers, and about a third did not specify how they implemented FL.}
%     \label{fig:fl_libraries}
% \end{figure}

Only 10 (27.1\%) articles included in this survey used established FL libraries, whereas 14 (37.8\%) developed their own libraries, and 13 (35.1\%) did not specify how the network was implemented. Fig. \ref{fig:fl_libraries} shows the FL libraries used by the analyzed articles.

% \input{figures/pie-chart-data-types}

%---Adam---
% Regarding the type and modality of data, Figure \ref{fig:type_of_data} shows that a majority (40.54\%) of works experimented with real clinical data followed by a 16.2\% using real finance data. 15\% used widespread datasets typically oriented to model benchmarking, and the remaining works used a diversity of real data types. According to Fig. \ref{fig:modality_of_data}, the predominant modality is tabular data (51.3\%) followed by images (18.1\%) and time series (21.6\%).

\begin{figure}[t]
    \centering
    \includegraphics{tikz/barchart-reported-influence}
    \vspace{-10pt}
    \caption{Number of articles reporting and not reporting the influence of FL on XAI, grouped as a function of the amount of data used. None of the papers using less than 1,000 data points reported any influence, whereas 9 papers working with more than 1,000 data points reported an influence.}
    \label{fig:barchart-reported-influence}
\end{figure}

% \begin{figure}
%     \centering
%     \includesvg[width=\columnwidth]{images/amount_data_vs_reported_influence.svg}
%     \caption{Number of papers reporting and not reporting influence of FL on XAI, grouped as a function of the amount of data used. None of the papers using less than 1,000 data points reported any influence, whereas 9 papers working with more than 1,000 data points reported an influence.}
%     \label{fig:influence_vs_data}
% \end{figure}

We examined the relation between the number of data points used in experiments and the reported influence between FL and XAI. Remarkably, no studies using less than 1,000 data points reported any influence, whereas 9 studies working with more than 1,000 data points reported an influence. Fig. \ref{fig:barchart-reported-influence} details this relation.

\begin{figure}[t]
    \centering
    \includegraphics{tikz/barchart-number-of-data-centers}
    \vspace{-15pt}
    \caption{Number of studies using real, simulated, or not specified FL setups, grouped as a function of the number of data centers in the FL network. Most cases using real FL setups used 3 data centers or fewer.}
    \label{fig:barchart-number-of-data-centers}
\end{figure}

% \begin{figure}
%     \centering
%     \includegraphics[width=\columnwidth]{figures/Number of data centers (2).png}
%     \caption{Number of papers using real, simulated, or not specified FL setups, grouped as a function of the number of data centers in the FL network. Most cases using real FL setups used 3 data centers or fewer.}%Florian
%     \label{fig:barchart-number-of-data-centers}
% \end{figure}
\begin{figure}[t]
    \centering
    \includegraphics{tikz/barchart-amount-of-data-user-in-each-paper}
    \vspace{-15pt}
    \caption{Number of studies using real, simulated, or not specified FL setups, grouped as a function of the number of data points. No papers in this survey reported real FL setups using more than 1,000,000 data points.}
    \label{fig:barchart-amount-of-data-user-in-each-paper}
\end{figure}

% \begin{figure}
%     \centering
%     \includegraphics[width=\columnwidth]{images/Amount of data used in each paper (1).png}
%     \caption{Number of papers using real, simulated, or not specified FL setups, grouped as a function of the number of data points. No papers in this survey reported real FL setups using more than 1,000,000 data points.}%Florian
%     \label{fig:barchart-amount-of-data-user-in-each-paper}
% \end{figure}

No articles in this survey reported real FL setups using more than 1,000,000 data points. Out of the 7 FL setups using between 100,000 and 1,000,000 data points, 3 used a real FL setup~\cite{p108_chen2022training,p48_chen2021fed,p34_razaa2022ecg}. Fig. \ref{fig:barchart-amount-of-data-user-in-each-paper} shows the distribution of articles using different amounts of data for the experiments, while the shading indicates whether the FL setup was real, simulated, or not specified.

The highest number of nodes (1,000) was simulated by \cite{p84_hou2021mitigating} using standard benchmark data (MNIST, CIFAR10), followed by \cite{liu2021learning} with 100 nodes, using text datasets for language modeling;~\cite{p46_haffar2023explaining} with 100 nodes, using physical activity and census data from the UCI ML repository; and \cite{p17_dong2022interpretable} with 66 nodes, using data from a cyberattack classification task. A minority of 9 out of 37 articles (24.3\%) used real setups, where \cite{p72_roschewitz2021ifedavg} used 12 nodes and all other studies implementing a real FL setup used 10 or fewer nodes. Eleven studies used between 3 and 10 nodes, 2 of which used a real setup, whereas 9 studies used 3 or fewer centers, 5 of which used a real setup. Fig. \ref{fig:barchart-number-of-data-centers} shows the distribution of articles using different amounts of FL centers, while the colors indicate whether the FL setup was real, simulated, or not specified.

%% FIELD OF APPLICATION; DATA TYPE

\begin{figure}[!t]
    \centering
    \includegraphics[width=\columnwidth]{tikz/pie-chart-data-modalities}
    \caption{Distribution of reviewed studies across the different modalities of data. Tabular and time-series data are predominant, images are used in about a fifth of the studies, and a minority focused on text data.}
    \label{fig:modality_of_data}
\end{figure}

% \begin{figure}
%     \centering
%     \includegraphics[width=\columnwidth]{images/Count of modality of data (simplified).png}
%     \caption{Distribution of reviewed papers across the different modalities of data. Tabular and time-series data are predominant, images are used in about a fifth of the studies, and a minority focused on text data.}%Adam
%     \label{fig:modality_of_data}
% \end{figure}
\begin{figure}[!t]
    \centering
    \includegraphics[width=\columnwidth]{tikz/pie-chart-field-distribution}
    \caption{Distribution of reviewed articles across different fields of application. Medicine and Life Sciences were predominant, followed by Finance and Cyber-security}\label{fig:field}
\end{figure}

% \begin{figure}
%     \centering
%     \includegraphics[width=\columnwidth]{images/field_application.png}
%     \caption{Distribution of reviewed papers across different fields of application. Medicine and Life Sciences were predominant, followed by Finance and Cyber-security.}%Aditya
%     \label{fig:field}
% \end{figure}

Fig. \ref{fig:field} shows the proportion of articles across the different fields of application, where \emph{medicine and life sciences} are predominant (40.5\%), followed by \emph{finance} (16.2\%), \emph{cybersecurity} (13.5\%), \emph{telecommunication} (8.1\%) and \emph{optimization} in other engineering fields (8.1\%) such as electricity, mechanics, and transportation. A small proportion (13.5\%) of the analyzed studies focused on the theoretical side of the ML techniques without mentioning any specific application. According to Fig. \ref{fig:modality_of_data}, the predominant modality is tabular data (51.3\%) followed by images (18.1\%) and time series (21.6\%).

\subsection{Interplay between FL and XAI}\label{s:interplay}
This section explores studies that have either integrated explainability into FL methods or analyzed how one technology influences the other. 
We first discuss instances where FL has impacted post-hoc explanations (Sec. \ref{ss:fl_explanations}). Next, we examine studies where explanation methods have influenced FL training as a design step in the FL process (Sec. \ref{ss:explanations_fl}). Following, we cover studies where FL has altered model interpretability (Sec. \ref{ss:fl_interpretability}). Finally, we discuss a study where the implementation of interpretable models has influenced FL training (Sec. \ref{ss:interpretability_fl}).

\subsubsection{FL Impacting Explanations}\label{ss:fl_explanations}
Among all articles reviewed, only \cite{p257_fiosina2021interpretable} quantitatively analyzed the impact of FL on explanations. Additionally, \cite{p206_bensaad2022trust} highlights privacy concerns associated with an explanation method used on FL models.

FL is applied to taxi travel-time prediction in \cite{p257_fiosina2021interpretable}, using time series and geographical data while maintaining data privacy. The study focuses on predicting taxi trip durations in the Brunswick region with a federated deep learning model, showing that FL can achieve accuracy similar to a centralized model when synchronization is optimized. To reduce communication overhead, the authors propose a method that lowers the frequency of model updates without significantly affecting performance. They evaluate explainability in FL using attribution methods such as DeepLIFT~\cite{ancona2017towards}, Saliency~\cite{simonyan2013deep}, Input X Gradient~\cite{shrikumar2016not}, Guided Backpropagation~\cite{springenberg2014striving}, Deconvolution~\cite{zeiler2014visualizing}, and Layer-wise Relevance Propagation~\cite{bach2015pixel}. Their results show that while local models in FL produce different variable importance rankings, aggregated feature attributions remain consistent with centralized models. This suggests that FL can incorporate explainability methods without compromising data privacy.

An FL model is trained in \cite{p206_bensaad2022trust} to predict the latency in the creation of a network slice in a mobile communications scenario. Each slice manager provides data regarding CPU/RAM capacity and usage, serving as an FL node. The models are evaluated both locally and globally using SHAP~\cite{lundberg_unified_2017}, LIME~\cite{ribeiro_why_2016}, Partial Dependence Plots (PDP)~\cite[Ch. 8.1]{molnar2020interpretable}, and RuleFit~\cite{friedman2008predictive}. For the PDP technique, the FL clients only share plots that display the percentage of feature impact on the target label to preserve data privacy. The influence is analyzed qualitatively with a focus on privacy concerns. Overall, it is shown that PDP explanations raise privacy concerns since they are executed on the client side.

% Potentially interesting source to add: Bogdanova et al. (July 2023) 
% https://link.springer.com/article/10.1007/s44230-023-00032-4

\subsubsection{Explanation Method Impacting FL Training}\label{ss:explanations_fl}
Among the articles that reported that the use of an explanation method impacts the FL training, a salient subset uses XAI to defend the FL process from the negative effects of defaulting nodes \cite{p106_malandrino2021node}, malicious behavior from certain FL nodes \cite{p84_hou2021mitigating,p46_haffar2023explaining}, and instances of the GAN attack in FL \cite{p263_ma2023research}. Other positive effects reported include improved accuracy in \cite{p1_younis2023flames,p15_yuan2022efficient,p16_demertzis2022explainable} and enhanced learning efficiency~\cite{p86_wang2022multi}. It must be noted that such benefits stem from incorporating XAI as a component of the FL algorithm design.

To detect malicious attacks on FL operations, \cite{p46_haffar2023explaining} uses a random forest (RF) to identify features causing incorrect predictions. Each client trains both a DL and an RF model on their training data. For samples that are misclassified, it calculates the feature importance of each feature regarding incorrect classification from all decision trees and using LIME~\cite{ribeiro_why_2016}. The change in a feature's importance over time is used to assign the contribution of each feature in misclassifying the data. Such an importance value provides insights into the level of influence of each log key in a sequence during an attack, thereby indicating which features should be most protected.

A novel FL protocol was proposed in \cite{p15_yuan2022efficient}, where a subset of all centers available online was selected to participate in each FL round, using Shapley values (computed using SHAP) as a heuristic to estimate FL contributions from each client. More specifically, at each round, the difference between local (feature-wise) and global aggregated Shapley values at each node is used to select participating clients in an FL process. The article reports that the proposed method improves both the efficiency and accuracy compared to the FedAvg protocol \cite{mcmahan2017communication}, which does not consider client contributions.

The idea in \cite{p84_hou2021mitigating} is to use explanation methods (specifically Backpropagation, Guided Backpropagation, DeepLIFT, GradCAM~\cite{selvaraju_grad-cam_2020}, and Integrated Gradients~\cite{sundararajan2017axiomatic}) to detect whether each client is using malicious data to enact a backdoor attack. To this end, so-called detection filters are developed. These consist of a classifier and an explanation method that respectively identify a likely backdoor attack and triggering features in the input data. The effectiveness of various explanation methods with different classifiers is tested, strengthening the FL process against backdoor attacks. Upon testing the detection accuracy in the presence of variable proportions of backdoor attacks, the proposed methodology proves the usefulness of the different explanation methods for backdoor attack prevention.

The technique proposed in \cite{p16_demertzis2022explainable} uses Shapley values and the Lipschitz constant to generate both local (feature-wise) and global explanations. Local Shapley values are compared with global Shapley values to refine the training of the local model, ensuring that only necessary characteristics are retrained, which allows for the personalization of the FL model for each user so that only the necessary characteristics of the model are retrained. A rigorous methodology is applied, resulting in increased accuracy, quantified explanations, and reduced dependence on input shift.

In an image classification task, Shapley values calculated via SHAP are used in \cite{p263_ma2023research} to identify the most important pixels for each local FL model and mask the pixels with the highest SHAP scores. The resulting dataset is then used for training the FL model. This method is proposed to protect the FL setup from adversarial attacks, specifically poisoning GAN attacks. By masking the majority of influential pixels in the input images, the difficulty of the GAN attack on FL is increased (although the exact influence of the method is not quantitatively assessed).

The algorithm proposed in \cite{p1_younis2023flames} is designed to explain the output of a time-series classifier. It extracts input subsequences that highly activate a time-domain convolutional neural network, facilitating their visualization. A graph capturing temporal dependencies is computed at each learning node. The central server aggregates these graphs into a global temporal evolution graph. By applying this method, an improved classification accuracy is claimed, compared to other FL algorithms such as FedAvg, FedRep, and FetchSGD.

The method proposed in \cite{p106_malandrino2021node}, termed Node Liability in Federated Learning (NL-FL), traces back ML decisions to training data sources in distributed settings. This method allows for the identification of misbehaving (defaulting) nodes that can be excluded from the training process. The influence of each node is quantified by measuring the classification accuracy after removing misbehaving nodes. The proposed method results in an improved prediction accuracy.

Adaptive sparse deep networks are implemented in \cite{p86_wang2022multi}, where parameters are shared via a multi-level federated network. At each round, weights are shared at the ``top'' and ``second'' sharing levels of the FL architecture, depending on the relevance values of the network calculated through Layerwise Relevance Propagation (LRP)~\cite{Bach2015lrp}. This approach provides good diagnostic results even when the FL dataset exhibits a non-independent and non-identically distributed (non-IID) structure.

\subsubsection{FL Impacting Model Interpretability}
\label{ss:fl_interpretability}

In this section, we discuss the impact that FL training has on the interpretability of the resulting models. The featured studies, \cite{p76_parra2022log} and \cite{p14_barcena2022approach}, discuss such an impact and its associated trade-offs.

An aggregation of client-based attention weights is investigated in \cite{p76_parra2022log} for a threat detection task in a cloud scenario. Using system logs as input data, each client predicts cyberattacks and computes local attention weights, which are claimed to enhance interpretability. The central server subsequently aggregates these attention weights to build a saliency map that provides insights into the impact of the different log keys on threat prediction. The influence of FL on model interpretability is assessed by comparing attention-based insights across three levels: individual cyberattacks at each client, individual attacks in the aggregated models, and aggregated attacks in the aggregated model. The aggregated insights proved to be more general but at the expense of reduced interpretability.

In \cite{p14_barcena2022approach}, a fuzzy rule-based system (FRBS) \cite{zhu2021horizontal} is trained in federation via a one-shot communication scheme. Each data silo computes its own FRBS, and the individual models are merged by the central server. The proposed FRBS uses a maximum-matching inference rule so that the inferred regression function is piecewise linear, which is inherently explainable. It is reported that FL impacts interpretability because the average number of model rules is higher when trained in the federated setting (vs. centralized). Following the intuitive idea that higher complexity implies lower interpretability, this suggests that training via FL caused lower interpretability.

\subsubsection{Interpretability Impacting FL Training}
\label{ss:interpretability_fl}

Only one of the reviewed articles discussed the effects of employing interpretable models on the process of training models in federation.
In the FRBS presented by \cite{p14_barcena2022approach}, the inherent interpretability of the employed Takagi-Sugeno-Kang FRBS~\cite{takagi1985fuzzy} models allows the central server to identify conflicts between rules inferred by data centers. This approach facilitates the reconciliation of such discrepancies, thereby enhancing the FL process.

\section{Discussion}
\label{s:discussion}

%In this section, we answer to the research questions %formulated in Sec. \ref{RQs} using the information presented in Sections \ref{s:summaries}, \ref{s:interplay}, and \ref{s:results}. Next, we 
%and provide an interpretation of the quantitative results charted in Sec. \ref{s:results}. 

\cmt{Interpretation of charted results}

%Regarding interpretation of the results, as observed in figures \ref{fig:field} and \ref{fig:type_of_data}, 
\scmt{Area of application}A majority of the reviewed studies focus on healthcare, finance, and engineering applications such as networking. %highlights a high degree of interdisciplinary research and suggests cross-fertilization among fields. This 
This contrasts the lack of studies in other high-stakes applications, such as social networks, language models, and supply-chain management, where user privacy and transparent decision-making are also crucial. The need to define explicability and privacy requirements usually originates from the end users' perspective, where data interoperability and reasoning behind automated decisions are key. The implementation of enabling technologies originates research toward using XAI for more technical goals, such as improving model accuracy, assessing the training process, and preventing malicious behavior, as observed in Sec. \ref{ss:explanations_fl}.

\scmt{FL types}A preference for HFL is evident among the studies surveyed, while combinations of FL types and applications of TFL remain relatively uncommon. This aligns with the observed higher prevalence of HFL in the literature compared to VFL~\cite{meng2023fedemb, cheung2022vertical}. Additionally, most studies implement cross-silo FL involving a small set of data centers. %[cf. fig. \ref{fig:number_data_centers}]. 
% Especially papers in real-world FL settings used a limited number of silos. 
% %(5 out of 9 papers with only 3 centers).
Whether this is due to limited data access or this is an accurate representation of FL networks remains unclear.

\scmt{Amount of data}
Most studies used simulated FL networks, and the ones done with a real FL setup used datasets with less than one million data samples. This raises questions of the cause-effect relation: while data sparsity has been named as a key driver for FL \cite{rank2024inter}, increasingly better-performing models are trained on billions of data points. 

\scmt{FL libraries $\rightarrow$ reporting guidelines}Also surprisingly, only a minority of the reviewed studies use established FL libraries. The remaining studies either develop their own libraries or lack descriptions of how the network was implemented. The absence of standardized reporting for used libraries can cause misleading observations if there are flaws in the libraries that are not identified during the analysis. 
A lack of transparency can hinder the development of standardized solutions, as inconsistencies in the reporting and usage of libraries complicate replicability. We also observed that many articles could improve their way of reporting data characteristics. Strict adherence to reporting guidelines such as TRIPOD+AI~\cite{Collinse078378} or MINIMAR~\cite{10.1093/jamia/ocaa088} would improve reproducibility and transparency. This can help standardize FL implementations, potentially accelerating the adoption of best practices in FL and XAI.

\scmt{Nomenclature} In the same line, not providing a definition of explainability [cf. Table \ref{tab:xai_definition}] can be a hurdle for reproducibility and further analysis. Works often disagree on what should be referred to as interpretability or an explanation method, which we aim to align in this study. Clarity and consensus in nomenclature are of utmost importance because they foster consistency in the evaluation of the understandability of ML results, improve communication among researchers and policymakers, and enable the development and reproducibility of new XAI methods.

\begin{table*}[t]
\centering
\label{tab:overview}
\begin{tabular}{|p{4.45cm}|p{5.9cm}|p{6.15cm}|}
\hline
\textbf{Interpretable Models Trainable in FL} & \textbf{Explanation Methods Affected by FL Training} & \textbf{XAI Techniques Integrated into FL Schemes} \\
\hline
\begin{list}{$\bullet$}{
  \setlength{\leftmargin}{1em}%  <-- left indentation
  \setlength{\labelwidth}{0.4em}%  <-- space for label
  \setlength{\labelsep}{0.5em}%    <-- gap between label and text
  \setlength{\itemsep}{0.4em}%     <-- vertical spacing between items
  \setlength{\topsep}{0.2em}%      <-- space before first item
}
  \item Fuzzy rule-based systems \cite{p14_barcena2022approach}
  \item Sparse SVMs \cite{p56_brisimi2018federated}
  \item Gradient Boosted Trees \cite{p17_dong2022interpretable}
  \item Cox regression models \cite{p49_masciocchi2022federated}
  \item Rule-based models via boosting \cite{p278_sokolovska2021vanishing}
\end{list} 
& 
\begin{list}{$\bullet$}{
  \setlength{\leftmargin}{1em}%  <-- left indentation
  \setlength{\labelwidth}{0.4em}%  <-- space for label
  \setlength{\labelsep}{0.5em}%    <-- gap between label and text
  \setlength{\itemsep}{0.4em}%     <-- vertical spacing between items
  \setlength{\topsep}{0.2em}%      <-- space before first item
}
  \item Feature relevance via SHAP, LRP \cite{p257_fiosina2021interpretable, p16_demertzis2022explainable}
  \item Saliency-based heatmaps (e.g., GradCAM) \cite{p84_hou2021mitigating}
  \item Attention weights aggregated across clients \cite{p76_parra2022log}
  \item Local explanation differences (central vs. FL) \cite{p257_fiosina2021interpretable}
  \item Privacy concerns in client-side explanations \cite{p206_bensaad2022trust}
\end{list}
& 
\begin{list}{$\bullet$}{
  \setlength{\leftmargin}{1em}%  <-- left indentation
  \setlength{\labelwidth}{0.4em}%  <-- space for label
  \setlength{\labelsep}{0.5em}%    <-- gap between label and text
  \setlength{\itemsep}{0.4em}%     <-- vertical spacing between items
  \setlength{\topsep}{0.2em}%      <-- space before first item
}
  \item Adversary detection by local explanation filters \cite{p84_hou2021mitigating}
  \item FL client selection via SHAP-based heuristics \cite{p15_yuan2022efficient}
  \item Personalized FL guided by explanation scores \cite{p16_demertzis2022explainable}
  \item XAI-informed retraining of model components \cite{p16_demertzis2022explainable}
  \item Feature-level privacy-preserving explanations \cite{p206_bensaad2022trust}
\end{list} 
\\
\hline
\end{tabular}
\caption{Overview of main forms of FL-XAI interplay}
\end{table*}

The remainder of this section discusses the implications of our findings regarding the interplay between FL and XAI in depth. A comprehensive summary is shown in Table \ref{tab:overview}.

\subsection{FL and Explanation Methods}
\scmt{FL + types of explanation}
Most of the reviewed articles employed post-hoc explanation methods, with a predominant focus on feature importance and local explanations.
%reflecting the prevalence of such methods in the literature.
None of the reviewed articles have proposed an ad-hoc FL method tailored to the ulterior application of explanation methods, highlighting the need for future approaches in that domain. Explanation methods designed for federated-trained ML models were limited to feature aggregation \cite{p257_fiosina2021interpretable}, control of information sharing \cite{p206_bensaad2022trust}, and counterfactual explanations in VFL \cite{chen2022explainableVFL}.

\scmt{Federated implementations of explanation methods}
Despite the prevalence of SHAP among the included studies, we did not find any mention of a potential federated implementation of SHAP when the supporting dataset is distributed among several data centers. Such a contribution would help exploit the representativeness of diverse data centers, potentially helping generate more accurate Shapley values. Among the studies using small amounts of data, very few used this easy-to-measure explanation, suggesting that an insufficient amount of data makes these observations challenging. Supporting data should not be part of the training dataset, so an insufficient number of data points likely hampers the generation of robust explanations, making it difficult to draw reliable conclusions.  
%---Luis---

% From discussion local vs global models and explanations
\scmt{Local vs Global (in FL and explanations)}
An important conceptual aspect that has received little attention in the reviewed literature is the distinction between local and global explanations in the context of FL. XAI refers to local explanations for individual instances and global explanations about model behavior as a whole. %Interestingly, both FL and XAI adopt the terms "local" and "global" but with different meanings: while FL discusses local models at client nodes and a global model at the server, XAI refers to local explanations for individual instances and global explanations about model behavior as a whole. The combination of FL and XAI demands careful disambiguation of these terms. Notably, whether FL affects local and global explanations differently remains an open question.
Global explanations may help assess whether FL results in a different model structure compared to centralized learning, potentially revealing divergences or biases in FL models. Conversely, local explanations, whether applied to client models or the aggregated model, may highlight instance-specific behaviors that differ due to local data distributions. Both levels of explanation may offer complementary insights, potentially revealing different aspects of the model’s behavior. For instance, a global explanation of a local model could help a client understand the model’s overall decision boundaries, while a local explanation of an aggregated model could clarify a prediction in a specific instance. Furthermore, structurally different models may yield similar global and local explanations if they implement equivalent decision boundaries, a phenomenon noted in neural networks due to overparameterization and symmetries~\cite{allen2019convergence}. Thus, whether FL leads to divergent or convergent explanations compared to centralized learning is an empirical question that remains vastly underexplored.

\scmt{Conclusion on FL + Explanation methods}
The scarcity of studies explicitly addressing and quantifying the influence of FL on explanations does not give grounds for concluding that there is no such influence, revealing an interesting research gap. Principled experimental work that objectively evaluates the impact of FL training on ML model structure is expected to clarify its implications for model explanations and interpretability. Future research should also aim to clarify under which conditions FL models align or deviate from their centralized counterparts in terms of both global and local explanations and how these explanations can guide model deployment in federated contexts.

\subsection {FL and Interpretation Methods}
\scmt{FL and interpretability types}The studies dealing with FL of interpretable models focused on algorithmic transparency. One proposed method learns a set of interpretable rules that reflect the structure of the FL network \cite{p14_barcena2022approach}. Fewer works deal with decomposability or simulatability, probably due to an increased difficulty in imposing those properties in an FL system. Interpretable models trained via FL methods were limited to fuzzy rule-based systems \cite{p14_barcena2022approach}, time series classifiers \cite{p65_liang2022fedtsc}, SVM \cite{p56_brisimi2018federated}, Cox proportional hazards \cite{p49_masciocchi2022federated}, sparse Bayesian models \cite{p246_kidd2022federated}, and decision trees \cite{p256_imakura2021interpretable}.
A particularly promising approach enables the training of federated classification models without relying on gradient descent-based methods \cite{p25_polato2022boosting}. This differs from most established methods, which focus on differentiable models.
The approach of \cite{p25_polato2022boosting} is based on adapting previously existing algorithms in the boosting family to FL. Other approaches proposed an optimization method to solve the sparse SVM problem in FL \cite{p56_brisimi2018federated}, a novel technique called Vanishing Boosted Weights  \cite{p278_sokolovska2021vanishing}, or an FL network based on gradient-boosting decision trees \cite{p17_dong2022interpretable}.

%Only the interpretable FL system proposed in \cite{p256_imakura2021interpretable} applied both horizontal and vertical FL, and \cite{chen2022explainableVFL} aimed to explain the outputs of a VFL system.
%\cite{p14_barcena2022approach,p65_liang2022fedtsc,p56_brisimi2018federated,p49_masciocchi2022federated,p246_kidd2022federated,p17_dong2022interpretable}. The methods in \cite{p257_fiosina2021interpretable,p206_bensaad2022trust} explain the outputs of HFL systems. The method introduced in \cite{p212_li2023privacy} to balance privacy and interpretability works with horizontal FL with adaptive differential privacy (ADP).
%
%
%\cmt{Under which conditions does training a model via FL affect a) its interpretability? b) the obtained explanations?}
%
\scmt{FL affecting interpretability}In general, experiments on the effect of FL on interpretability were very sparse. We identified only two articles observing an influence of FL on model interpretability \cite{p14_barcena2022approach,p76_parra2022log}.

\scmt{FL, interpretability, local vs global metrics}In \cite{p76_parra2022log} it was reported that aggregating interpretability metrics from different nodes led to more general but less interpretable global insights. This fact supports the idea that FL produces better generalizing models at the cost of missing some interpretable insights from the local nodes of the FL framework. Also, aggregating feature relevance or attention weights across nodes can yield more generalized global insights but can also result in sacrificing some degree of localized interpretability. This was reported in the case of an attention mechanism \cite{p76_parra2022log} where each client trained a local model on system log data and generated local attention weights. The global saliency map resulting from aggregating the local attention weights from the different nodes, provided a global saliency map. The contribution of individual client models in such a map was diluted, possibly leading to unique local patterns getting lost during global aggregation. The FRBS applied by \cite{p14_barcena2022approach} in a federated context involved rule generation at each data silo and merging them at the central server. Experiments observed an increased number of rules in the global model compared to local ones, indicating a growth in model complexity upon aggregating the local models, yielding a more expressive global model that is less interpretable than individual local models. 
These results suggest that, while FL can improve overall model performance and integrate information from diverse data sources, outputs from the global FL model could become more difficult to interpret than those from local models (trained using data from single sites). This may negatively affect applications where the transparency of the local models is beneficial. One example is medical prognosis, where each local model is associated with a clinical site or hospital, and the distributions of disease features vary across hospitals. This issue could benefit from new aggregation strategies or federated explanation techniques that retain node-specific insights without compromising global model integrity.

\cmt{Limitations of our study}
A limitation of this study and suggestion for future work is that the FL algorithm (e.g. FedAvg) used in each study was not extracted, which could be a relevant factor in analyzing the impact on explanations. As stated above, adherence to information reporting is important, and future reviews on the field should look closer at this aspect.
Further, we conducted our search in 2023. In the ever-changing domain of ML model development, further developments could have been achieved in the meantime.

\section{Conclusion}\label{s:conclusion}
This review studies the interplay between FL and XAI by mapping methodological and experimental contributions. We identified a research gap in which few studies quantify the impact of FL on explanations. Moreover, the impact of FL on model interpretability and explanations remains unclear, revealing the need for studies that quantify such impact. There is a need for rigorous experimental and analytical research to assess how FL training influences the structure of an ML model and its implications for explainability and interpretability. Additional research should also provide guidance for practitioners on the responsible implementation of FL with XAI, particularly in critical application fields such as healthcare, finance, and engineering. This guidance will help ensure the responsible deployment of AI systems that balance model performance with transparency.

Many articles do not use consistent terminology for explainability, which complicates the analysis. It is important for future research to clearly define terms and use standardized nomenclature. Another important finding is that a minority of the studies specify which FL libraries were used. Future publications should include details on the employed libraries or provide source code for their custom implementations. Furthermore, the lack of adequate reporting of data characteristics calls for strict adherence to reporting guidelines to ensure reproducibility, transparency, and auditability. We underscore the need for more structured and transparent research practices at the intersection of FL and XAI. Establishing clear definitions and consistent methodologies will be key in advancing the field and addressing the identified gaps. As demand for FL and XAI continues to grow, particularly in high-stakes application fields such as medicine, the importance of rigorous, transparent, and reproducible research cannot be overstated.

Moreover, while local explanations are frequently applied in FL contexts, no dedicated federated implementations of local explanation methods exist. Aggregating local explanations to obtain global insights often dilutes client-specific patterns, and sharing local explanations may raise privacy concerns. Developing federated-specific local explanation techniques remains an open research challenge.

\bibliography{references} 

\clearpage

\appendices

\section{Extended List of Reviewed Papers}
\label{appendix:summaries}

In this section, we describe how each selected paper deals with FL and XAI in combination. The first part of the section discusses works where FL is combined with explanation methods, both methodological (Sec. \ref{ss:fl_explanations_methodological}) and applied (Sec. \ref{ss:fl_explanations_applied}). The second part of the section discusses papers combining FL and interpretable models, both methodological (Sec. \ref{ss:fl_interpretable_methodological}) and applied (Sec. \ref{ss:fl_interpretable_applied}).

\subsection{Combining FL and Explanation Methods: Methodological Contributions}
\label{ss:fl_explanations_methodological}

%Luis, Adam
\cite{p15_yuan2022efficient} proposes a novel FL protocol where a subset of all centers available online are selected to participate in each FL round by calculating the difference between the aggregated global and local feature contribution (using SHAP~\cite{lundberg_unified_2017}). A more efficient and improved performance is shown when using this selection.

%Florian, Adam
\cite{p41_chen2022explainable} proposes the first counterfactual explanation method for VFL. They show the validity by retraining VFL models on banking data while leaving out a varying number of features concerning their counterfactual importance rate and comparing them against a random selection of variables.

%Aditya, Adam
\cite{p84_hou2021mitigating} uses explanation methods such as GradCAM~\cite{selvaraju_grad-cam_2020} to detect whether each participant is using malicious data to enact a backdoor attack %by using techniques such as label flipping. 
by developing so-called detection filters. 
These consist of a classifier and an explanation method that identify a likely backdoor attack and triggering features in the input data, respectively.
The effectiveness of various explanation methods with different classifiers is tested, strengthening the FL process against backdoor attacks.
%xAI is used to detect trigger features of adversarial attacks on the learning process. The XAI doesn't explain the contribution of each client in the training process in any other constructive way.

%Luis, Aditya
\cite{p263_ma2023research} tackles an image classification task by calculating Shapley values by SHAP and using them to find the most important pixels for every local FL model and mask the pixels with the highest SHAP score. The resulting dataset is used for training the FL model. This method is proposed to protect the FL setup from adversarial attacks, specifically poisoning GAN attacks.

%Aditya, Florian
\cite{p46_haffar2023explaining} uses random forest (RF) algorithms to detect features causing wrong predictions, with an emphasis on detecting malicious attacks on the FL operation. Each participant trains a DL and RF model on its training data. For the samples that are wrongly classified, it calculates the average feature importance of all decision trees with LIME~\cite{ribeiro_why_2016} that will result in the same wrong classification. It uses the change in this feature's importance over time to predict the contribution of each feature in wrongly classifying the data.

%Tobias, Luis
\cite{p1_younis2023flames} proposes and designs an algorithm for explaining the output of a time-series classifier. It extracts and visualizes the input subsequences that highly activate a convolutional neural network. A graph capturing temporal dependencies is computed at each learning node. The central server aggregates the obtained graphs into a global temporal evolution graph.

%Aditya, Steven
\cite{p99_kang2022privacy} introduces PrADA, a privacy-preserving federated adversarial domain adaptation technique addressing cross-silo federated domain adaptation issues. PrADA mitigates sample and feature scarcity by employing VFL with a feature-rich party and implementing adversarial domain adaptation from a sample-abundant source. For interpretability, features are segregated into semantically meaningful groups for fine-grained adaptation based on Shapley values computed using SHAP.

%Luis, Steven
\cite{p106_malandrino2021node} proposes a method, namely node liability in federated learning (NL-FL), to trace back ML decisions to training data sources in distributed settings. The method allows for the identification of misbehaving nodes that can be excluded from the training process, resulting in improved prediction results.

%Luis, Florian
\cite{p86_wang2022multi} implements interpretable adaptive sparse deep networks that exchange NN parameters employing a multi-level federated network. Whether those weights are shared at the ``top sharing level'' of the FL architecture depends on the relevance values of the network calculated through layerwise relevance propagation (LRP). The approach provides good diagnostic results even when the FL dataset is under a non-independent identical distribution (NOIID).

\subsection{Combining FL and Explanation Methods: Applied Contributions}
\label{ss:fl_explanations_applied}

%Luis, Adam
\cite{p224_slazyk2022deep} trains FL models to segment lung X-ray images and detect signs of pneumonia. Grad-CAM is used to highlight parts of the images that contribute to a detection. It is concluded that a model trained on segmented images has less accuracy, but the pixels highlighted by Grad-CAM focus more on the lung area. It is also reported that training the model in the FL manner helps maintain generalizability and avoid overfitting. A fixed number of FL rounds and a greater number of local iterations result in more accuracy.

%Luis, Adam
\cite{p243_xu2023federated} aims to predict residential load using a recurrent neural network (RNN). To explain the importance of features, the authors propose a novel automatic relevance determination (ARD) method. An iterative federated clustering algorithm (IFCA) is used, which keeps several central models while clustering the input data sequences, and each model is updated using data in its associated cluster. No interaction between ARD and IFCA is reported.

%Florian, Adam
\cite{p257_fiosina2021interpretable} develops an FL procedure for taxi travel-time prediction based on time-series and geographical data. Authors develop a federated feature attribution aggregation method and test how similar the FL-calculated explanations are compared to central calculation. Many XAI techniques are tested, and all result in similarly low differences.

%Adam, Aditya
\cite{p16_demertzis2022explainable} compares the use of Shapley values and Lipschitz constant for generating both local and global explanations and uses this information to update the model. It allows for the personalization of the FL model for each user, so that only the necessary characteristics of the model are retrained based on the respective needs and the events it is called to respond to.

%Florian, Aditya
\cite{p58_huong2022federated} uses Shapley values computed via SHAP to explain the outputs of an FL model trained on edge devices to its operators. It takes the model and the test data as inputs to construct a local linear regression explanation model. Subsequently, the explanatory model computes the Shapley values of classified anomalies and displays them visually. As feature values are measured by sensors, the explanations help operators determine the sensors likely causing an abnormality and make a faster detection response.

%Luis, Florian
\cite{p206_bensaad2022trust} proposes to train an FL model to predict the latency in the creation of a network slice. Each slice manager provides data regarding CPU/RAM capacity and usage and serves as an FL node. The models are evaluated on a local and global level using SHAP, LIME, partial dependent plot (PDP)~\cite[Ch. 8.1]{molnar2020interpretable} and RuleFit~\cite{friedman2008predictive}. Overall, they show that PDP explanations raise privacy concerns since they are run on the client side.

%Steven, Florian
\cite{p269_sateeshambesange2023federated} discusses the use of FL and transfer learning to enhance AI-based lung segmentation. The study achieved good segmentation accuracy using local system data and pre-trained weights from U-net models. The approach utilized a reduced number of nodes with varying dataset sizes and incorporated a model-agnostic explanation method (activation map) to clarify the results.

%Tobias, Luis
\cite{p108_chen2022training} trains a language model that takes free text from electronic medical records and classifies a patient's disease into one of the International Classification of Diseases (ICD-10) codes. They showed explanations for the predictions by highlighting input words via a label attention architecture. FL is realized via Flower \cite{beutel2020flower} to integrate training data from 3 different sites while keeping data privacy.

%Steven, Luis
\cite{p209_raza2023anofed} introduces AnoFed, a framework integrating transformer-based Autoencoders (AEs) and Variational Autoencoders (VAEs) with Support Vector Data Description (SVDD) in a federated environment, specifically for ECG anomaly detection, optimizing computational efficiency. A combined design of the VAE/AE and SVDD incorporates kernel density estimation for adaptive anomaly detection. Moreover, it includes a module that explains the anomaly detection output by identifying key segments of the ECG signal that show the maximum reconstruction loss.

%Tobias, Steven
\cite{p34_razaa2022ecg} presents an ECG-based arrhythmia classification framework that trains convolutional DNNs via FL and includes an XAI module computing activation mappings in the ECG signal utilizing GradCAM. The framework addresses data availability, privacy, and interpretability challenges.

%Florian, Adam
\cite{p205rahman2021secure} proposes a framework for FL in connected medical devices with blockchain integration for safe model parameter exchange. That framework is showcased by a hybrid implementation of actual hardware nodes and simulated distribution of publicly available datasets in many use cases. Explanations are shown in two cases, however, the results and impact of FL are not discussed.

\subsection{Papers combining FL and Interpretable Models: Methodological Contributions}
\label{ss:fl_interpretable_methodological}

%Aditya, Luis
\cite{p25_polato2022boosting} proposes an FL algorithm that builds federated classification models without relying on gradient descent-based methods. Therefore, the class of algorithms that can be learned via FL is not restricted to models whose output is differentiable concerning the model parameters. Based on AdaBoost~\cite{ying2013advance}, it effectively combines gradient-free classifiers, which may be learned independently by the FL clients.

%Luis, Inga
\cite{p14_barcena2022approach} trains a fuzzy rule-based system (FRBS) \cite{zhu2021horizontal} in federation via a one-shot communication scheme %(first-order Takagi- Sugeno-Kang FRBS (TSK-FRBS))
where each data silo computes their own FRBS and the individual models are merged by the central server. The proposed FRBS uses a maximum-matching inference rule, so the inferred regression function is piecewise linear, which is inherently explainable.

%Aditya, Luis
\cite{p65_liang2022fedtsc} proposes an FL framework for time-series classification using interpretable, human-understandable time-series features, namely shapelet features, interval features, and dictionary features. The paper claims to guarantee interpretability for the learning-initiating party by ensuring that it can access the aforementioned features without data leakage. To ensure security, the solution incorporates secure feature extraction protocols, secure model training protocols, additive secret sharing schemes, and secure computation protocols.

%Florian, Steven
\cite{p33_pedrycz2021interpretability} highlights the importance of using information granules for better interpretability, focusing on unsupervised federated learning and enhancing rule-based models through granule decomposition and linguistic approximation.

%Luis, Steven
\cite{p48_chen2021fed} proposes an Efficient and Interpretable Inference Framework for Decision Tree Ensembles in FL (Fed-EINI), based on streamlined multi-party communication. The paper highlights the challenge of current privacy-preserving ML frameworks compromising model interpretability to prevent data breaches. The proposed solution enhances interpretability by disclosing feature meanings while maintaining privacy.

%Aditya, Steven
\cite{p72_roschewitz2021ifedavg} presents an interpretable data interoperability method for FL called iFedAvg to address the low interoperability due to client data inconsistencies, among other challenges. The iFedAvg method uses personalized layers to adjust for local data shifts, like age differences, directly within input features, which maintains privacy while allowing for direct interpretability. The difference in values of private weight and bias of the input layer of each participant captures the inherent shift in data. It was tested on public benchmarks and a large, real-world Ebola dataset.

%Aditya, Steven
\cite{p254_liu2023federated} introduces a group personalization strategy in FL to address client drift in settings with distinct client partitions. The authors fine-tuned a global FL model with another FL process for each homogeneous client group and then adapted it per client. The method is tested on real-world language modeling datasets and aligns with Bayesian hierarchical modeling principles.

%Adam, Steven
\cite{p256_imakura2021interpretable} introduces an interpretable FL system for collaborative data analysis across distributed networks using interpretable models such as decision trees, sharing intermediate representations of the data rather than models. The result is an interpretable model that performs better than individual analyses and nearly as well as centralized methods.

%Steven, Vince
\cite{p212_li2023privacy} introduces adaptive differential privacy (ADP) in FL to balance privacy and model interpretability, assessed by inspecting Grad-CAM heatmaps. ADP selectively injects noise into client model gradients, mitigating gradient leakage attacks while preserving interpretability. Through theoretical and experimental analyses on IID and NOIID data, it overcomes the limitations of traditional differential privacy, demonstrating a harmonious blend of data privacy safeguards and interpretability.

%Luis, Adam
\cite{p56_brisimi2018federated} proposes a framework for solving sparse support vector machine (SVM) classification in a distributed fashion. Its performance was demonstrated on an electronic health record (EHR) dataset. The proposed algorithm has an improved convergence rate compared to several alternatives. Interpretability is assessed by achieving a classifier with fewer features considered as highly important for predictions.

\subsection{Papers Combining FL and Interpretable Models: Applied Contributions}
\label{ss:fl_interpretable_applied}

%Luis, Florian
\cite{p17_dong2022interpretable} implements an FL network based on Gradient Boosting Decision Trees (GBDT). These GBDT are considered transparent models, and therefore, their FL network is considered transparent as well. They apply their algorithm on network intrusion data sets and claim that the model is interpretable for human experts.

%Steven, Adam
\cite{p5_zheng2021federated} introduces %FL-LRBC, 
a %big data and AI-driven 
method in financial risk management for credit scoring and rating with big-data capabilities. Using a VFL framework allows multiple agencies to collaboratively train an optimized scorecard model. Performance is showcased on two finance datasets.

%Luis, Aditya
\cite{p278_sokolovska2021vanishing} proposes a novel technique called Vanishing Boosted Weights for fine-tuning models trained by a GBDT algorithm, along with an FL version of this approach. Based on stumps, the model remains interpretable due to their limited number. Iterative adjustment of each stump's output value occurs multiple times by incorporating a vanishing sequence of values.

%Luis, Florian
\cite{p76_parra2022log} investigates client-based attention weight aggregation in a threat-detection task within a cloud scenario using system logs as input data. Each client predicts local attentions, which are claimed to enhance interpretability,  and the central server subsequently aggregates the attention weights to build a saliency map that provides insights on the impact of the different log keys on the threat prediction.

%Tobias, Luis
\cite{p49_masciocchi2022federated} proposes an adaptation for FL of the Cox Proportional Hazards regression model with LASSO regularization. Such an estimator is used as a feature selector in the context of survival analysis and personalized medicine. Including LASSO regularization enacts a feature selection, contributing to the model's interpretability.

%Steven, Luis
\cite{p246_kidd2022federated} proposes methods for training sparse Bayesian models in federation, allowing pooling from multiple data sources without privacy issues and offering principled uncertainty quantification. The methods are based on Markov Chain Monte Carlo (MCMC) updating steps, where the order of updating steps can be interchanged so the communication between local servers and the global server can be reduced by running multiple local steps per global aggregation.

%Luis, Steven
\cite{p55_renda2022federated} explores the FL of interpretable models in 5G and 6G systems, focusing on automated vehicle networking. The approach addresses gaps in existing AI-based solutions for wireless networks, particularly in vehicle-to-everything (V2X) environments. The methodology offers decentralized, efficient intelligence, enhancing operational trustworthiness and data management.

\end{document}